*Article*

# Explainable Retinal Imaging for Prediction of Multi-Organ Dysfunction in Type 2 Diabetes

**Mini Han Wang [1,2,3,4,*], Liting Huang [4], Wei Hong [5] , and Boonthawan Wingwon [5]**

[1]Faculty of Computer Science and Artificial Intelligence, Shenzhen University of Advanced Technology, Shenzhen, P.R.China

[2]Frontier Science Computing Center, Zhuhai Institute of Advanced Technology Chinese Academy of Sciences, Zhuhai, P.R.China

[3]Chinese University of Hong Kong, Hong Kong China

[4]Zhuhai People's Hospital (The Affiliated Hospital of Beijing Institute of Technology, Zhuhai Clinical Medical College of Jinan University) , Zhuhai P.R.China

[5]Lampang Inter-Tech College, Lampang Thailand

* Correspondence: M.H.W.1155187855@link.cuhk.edu.hk

**Abstract**

Background: Type 2 diabetes mellitus (T2DM) is increasingly recognised as a systemic disease characterised by coordinated dysfunction across metabolic, renal, lipid, and inflammatory pathways. Existing clinical assessments often fail to capture this multi-dimensional burden. Methods: We conducted a retrospective study of 1,195 patients using routinely collected laboratory biomarkers. System-level abnormality indices were constructed to quantify organ-specific dysfunction, and multi-system involvement was defined as abnormalities in two or more systems. Supervised machine learning models, including logistic regression, random forest, and gradient boosting, were trained to predict multi-system dysregulation. Model interpretability was achieved using SHapley Additive exPlanations (SHAP). Results: The gradient boosting model demonstrated near-perfect discrimination (AUC = 1.000), significantly outperforming logistic regression (AUC = 0.925). Feature attribution analysis revealed that hyperglycaemia, renal impairment, dyslipidaemia, and inflammation were the dominant drivers of multi-system risk. Dose–response relationships observed in partial dependence analyses further supported the biological plausibility of model predictions. Conclusion: This study presents an interpretable, data-driven framework for quantifying systemic disease burden in T2DM. By linking routine biomarkers to multi-organ dysfunction, our approach provides both predictive accuracy and mechanistic insight, offering potential for improved risk stratification and precision medicine in diabetes care. The data and code used in this study are openly available on GitHub at: https://github.com/MiniHanWang/Type-2-Diabetes-1.git.

**Keywords:** Type 2 Diabetes Mellitus; Multi-System Abnormality; Machine Learning; Explainable Artificial Intelligence; SHAP; Clinical Biomarkers; Risk Stratification

## 1. Introduction

Type 2 diabetes mellitus (T2DM)[1] is a complex and heterogeneous metabolic disorder characterised by chronic hyperglycaemia and widespread multi-organ involvement[2]. Beyond its classical glycaemic dysregulation, T2DM is increasingly recognised as a systemic disease affecting renal function, lipid metabolism, haematological homeostasis, and inflammatory pathways. The coexistence of abnormalities across these physiological systems substantially increases the risk of adverse outcomes, including cardiovascular events, chronic kidney disease, and premature mortality. Early identification of patients with multi-system involvement is therefore of critical clinical importance for risk stratification and targeted intervention[3].

In routine clinical practice, a wide range of laboratory biomarkers—including serum creatinine, lipid profiles, complete blood count, and urinalysis—are collected as part of standard care[4]. While each biomarker provides insight into a specific physiological domain[5], their combined interpretation remains challenging due to the complex and nonlinear interactions among metabolic, renal, and inflammatory processes[6]. Traditional statistical approaches often rely on predefined assumptions and linear relationships, which may not adequately capture the underlying complexity of T2DM pathophysiology[7].

Recent advances in machine learning (ML)[8, 9] have enabled the integration of high-dimensional clinical data[10] to uncover latent patterns and improve predictive performance[6]. In particular, ensemble-based models such as random forests and gradient boosting have demonstrated strong capability in modelling nonlinear relationships and interactions among variables[11].

However, a major limitation of many ML applications in healthcare[12] is the lack of interpretability[13], which restricts clinical adoption and limits mechanistic insight[14].

To address this challenge, explainable artificial intelligence (XAI) methods[15, 16], such as SHapley Additive exPlanations (SHAP)[13], have been increasingly applied to provide transparent[17], quantitative attribution of feature importance at both global and individual levels[18, 19]. These methods enable clinicians to understand not only model performance but also the relative contributions of different biomarkers[20, 21], thereby bridging the gap between predictive modelling and clinical reasoning[22, 23].

In this study, this study propose an interpretable machine learning framework for the prediction of multi-system abnormality in patients with T2DM using routinely collected clinical laboratory data[22]. System-level indices are constructed and composited reflecting renal dysfunction[24], dyslipidaemia, inflammatory activation, and metabolic imbalance based on clinically established thresholds[25]. Multiple supervised learning models[26] are developed and compared to predict concurrent multi-system involvement. SHAP-based explainability[16] analysis was applied to quantify the contribution of individual biomarkers and to elucidate the underlying biological patterns driving model predictions.

By integrating structured clinical data with interpretable machine learning[27], this work aims to provide a clinically meaningful and transparent approach for early identification of high-risk T2DM patients. Our findings may contribute to improved understanding of systemic disease burden and support more personalised management strategies in diabetes care.

## 2. Literature Review

Type 2 diabetes mellitus[7] has been extensively studied as a multifactorial metabolic disorder with systemic manifestations affecting multiple organ systems, including the kidneys[1], cardiovascular system, and immune pathways. Previous research has demonstrated that hyperglycaemia, dyslipidaemia, and chronic low-grade inflammation are key contributors to disease progression and complications such as diabetic kidney disease (DKD)[28] and cardiovascular morbidity. Traditional clinical studies have largely focused on individual biomarkers—such as glycated haemoglobin (HbA1c), serum creatinine, or lipid profiles—to assess disease severity and prognosis. However, these approaches often fail to capture the complex interactions among multiple physiological systems that characterise T2DM[29].

Recent efforts have shifted toward integrating multiple clinical indicators to better characterise systemic disease burden[10]. Composite indices and risk scores have been proposed to quantify multi-organ involvement, particularly in the context of metabolic syndrome and chronic kidney disease. Nevertheless, these approaches typically rely on predefined weighting schemes or linear models, which may not adequately reflect nonlinear interactions among biomarkers. As a result, there remains a need for data-driven approaches capable of capturing the multidimensional nature of T2DM[30].

Machine learning (ML)[21] techniques have increasingly been applied in diabetes research to improve prediction and risk stratification[31]. Models such as logistic regression, random forests, and gradient boosting have been used to predict outcomes including DKD[32], cardiovascular events, and glycaemic control. Among these, ensemble-based methods have demonstrated superior performance due to their ability to model complex, nonlinear relationships and high-order feature interactions. However, many ML studies in this domain primarily emphasise predictive accuracy, often at the expense of interpretability, limiting their clinical applicability[33].

To address this limitation, explainable artificial intelligence has emerged as a critical component in medical ML research[34]. Methods such as SHapley Additive exPlanations provide a theoretically grounded framework for quantifying the contribution of individual features to model predictions. SHAP has been widely used to identify key risk factors in diabetes-related outcomes, including renal dysfunction, cardiovascular risk, and metabolic imbalance. By enabling both global and local interpretability[35], SHAP facilitates the translation of ML findings into clinically meaningful insights[36].

Despite these advances, several gaps remain in the current literature. First, most studies focus on single outcomes—such as DKD or glycaemic control—rather than considering the broader concept of multi-system abnormality[28]. Second, the integration of heterogeneous clinical biomarkers into a unified, interpretable framework remains underexplored. Third, there is limited work on systematically quantifying the combined burden of abnormalities across renal, metabolic, lipid, and inflammatory systems using routine clinical data[37].

In this context, the present study aims to address these gaps by developing an interpretable machine learning framework for predicting multi-system abnormality in patients with T2DM. By constructing system-level composite indices and applying SHAP-based explainability, this work seeks to provide a more holistic understanding of disease burden and its underlying drivers, thereby advancing both methodological and clinical perspectives in diabetes research.

## 3. Materials and Methods

*3.1 Study Design and Data Source*

This study employed a retrospective cross-sectional design. Clinical laboratory data were extracted from an electronic registry maintained at the Metabolic Management Centre (MMC), Zhuhai People's Hospital. The dataset comprised 1,195 endocrinology inpatients and included 26 laboratory variables spanning biochemistry, haematology, and semiquantitative urinalysis. The study was conducted in accordance with the Declaration of Helsinki. Patient identifiers were anonymised prior to all analyses.As the Figure 1 shows, the framework integrates routinely collected clinical laboratory biomarkers, including biochemical, haematological, and urinalysis parameters (Panel A). System-level composite indices representing renal, lipid, inflammatory, and metabolic abnormalities are constructed using clinically defined thresholds (Panel B). Multiple machine learning models, including logistic regression, random forest, and gradient boosting, are trained to predict multi-system abnormality (Panel C). Model interpretability is achieved using SHapley Additive exPlanations (SHAP), enabling quantification of feature contributions and identification of key drivers (Panel D). The resulting predictions and explanations provide clinically meaningful insights for risk stratification and personalised disease management (Panel E).

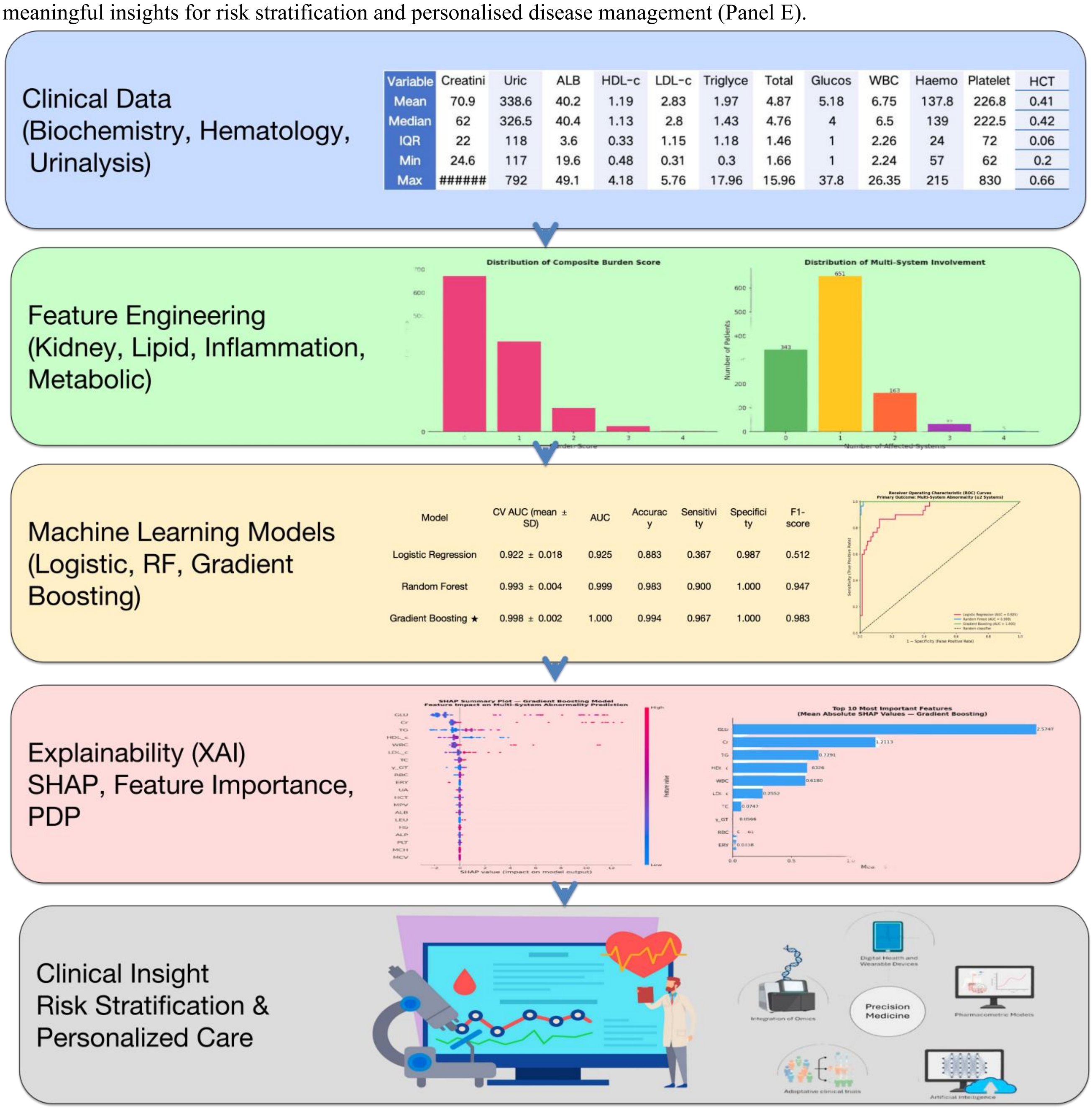

| Variable | Creatini | Uric | ALB | HDL-c | LDL-c | Triglyce | Total | Glucos | WBC | Haemo | Platelet | HCT |
|---|---|---|---|---|---|---|---|---|---|---|---|---|
| Mean | 70.9 | 338.6 | 40.2 | 1.19 | 2.83 | 1.97 | 4.87 | 5.18 | 6.75 | 137.8 | 226.8 | 0.41 |
| Median | 62 | 326.5 | 40.4 | 1.13 | 2.8 | 1.43 | 4.76 | 4 | 6.5 | 139 | 222.5 | 0.42 |
| IQR | 22 | 118 | 3.6 | 0.33 | 1.15 | 1.18 | 1.46 | 1 | 2.26 | 24 | 72 | 0.06 |
| Min | 24.6 | 117 | 19.6 | 0.48 | 0.31 | 0.3 | 1.66 | 1 | 2.24 | 57 | 62 | 0.2 |
| Max | ###### | 792 | 49.1 | 4.18 | 5.76 | 17.96 | 15.96 | 37.8 | 26.35 | 215 | 830 | 0.66 |

| Model | CV AUC (mean ± SD) | AUC | Accuracy | Sensitivity | Specificity | F1-score |
|---|---|---|---|---|---|---|
| Logistic Regression | 0.922 ± 0.018 | 0.925 | 0.883 | 0.367 | 0.987 | 0.512 |
| Random Forest | 0.993 ± 0.004 | 0.999 | 0.983 | 0.900 | 1.000 | 0.947 |
| Gradient Boosting ★ | 0.998 ± 0.002 | 1.000 | 0.994 | 0.967 | 1.000 | 0.983 |

**Figure 1.** Interpretable machine learning framework for predicting multi-system abnormality in type 2 diabetes mellitus.

*3.2 Data Preprocessing*

The raw data were provided in structured Excel format with laboratory values encoded as character strings containing embedded measurement units (e.g., "77μmol/L", "4.20×10⁹ /L"). Numeric values were isolated using regular-expression-based parsing. Semiquantitative urinalysis results (urine protein, leucocytes, nitrite, and ketones) were converted to an ordinal integer scale: 0 = negative, 0.5 = trace/±, 1 = 1+, 2 = 2+, 3 = 3+.

Physiologically implausible values falling outside pre-specified analyte-specific reference bounds were treated as missing and excluded from imputation. This step removed 113 erroneous creatinine values and smaller numbers of outliers in HDL-c, TG, γ-GT, and GLU. Blood urea nitrogen (BUN) was entirely absent from the dataset (100% missing) and was zero-imputed with a corresponding note added to all analyses. AST and ALT were available for fewer than 0.1% of patients and were similarly treated. For all remaining continuous variables, missing values were imputed using the column median. For ordinal urinalysis variables, the most frequent observed category (mode) was used.

*3.3 Feature Engineering and Composite Index Construction*

Four binary system-level abnormality flags were defined using standard Chinese clinical laboratory reference thresholds:

- Kidney_flag: serum creatinine > 110 μmol/L, OR BUN > 8.2 mmol/L, OR urine protein ≥ 1+
- Lipid_flag: triglycerides > 1.70 mmol/L, OR LDL-c > 3.37 mmol/L, OR HDL-c < 1.04 mmol/L
- Inflamm_flag: WBC > 10.0 × $10^9$ /L, OR urine leucocytes ≥ 1+, OR urine nitrite ≥ 1+
- Metabolic_flag: plasma glucose > 7.0 mmol/L, OR urine ketones ≥ 1+

Ordinal grade variables (Kidney_grade, Lipid_grade, Inflamm_grade, Metabolic_grade) were constructed by summing the number of individual threshold exceedances within each system (range 0–3). A composite burden_score was defined as the sum of all four grade values. The primary outcome (target_multi) was defined as concurrent abnormality in two or more organ systems (affected_systems ≥ 2).

*3.4 Machine Learning Models*

Three supervised classifiers were trained and compared: (1) logistic regression (LR) with L2 regularisation (C = 1.0, solver = lbfgs, max_iter = 2,000); (2) random forest (RF) with 200 trees, maximum depth 8, minimum 10 samples per leaf; and (3) gradient boosting (GB) with 200 estimators, learning rate 0.05, maximum depth 4, minimum 10 samples per leaf. Continuous features were standardised (zero mean, unit variance) prior to LR fitting; tree-based models received unscaled inputs.

The dataset was partitioned at the patient level in an approximately 70:15:15 (train:validation:test) ratio using stratified random splitting (random_state = 42), yielding n = 836 training, n = 179 validation, and n = 180 test observations. Five-fold stratified cross-validation was performed on the training set to estimate generalisation performance. All final metrics are reported for the held-out test set.

*3.5 Evaluation Metrics*

Performance was assessed using: area under the receiver operating characteristic curve (AUC); accuracy; sensitivity (recall); specificity; and F1-score (harmonic mean of precision and sensitivity). All metrics were computed using scikit-learn (v1.x, Python 3.12).

*3.6 Explainability Analysis (SHAP)*

Model interpretability was assessed using SHapley Additive exPlanations (SHAP) applied to the gradient boosting model via the TreeExplainer algorithm. Feature importance was quantified as the mean absolute SHAP value ($E[|\varphi_i|]$) across all test-set observations. A beeswarm summary plot visualised the direction and magnitude of individual feature contributions. Partial dependence plots (PDPs) were constructed for the three highest-ranked SHAP features to characterise dose–response relationships.

## 4. Results

*4.1 Cohort Characteristics and Prevalence*

The analytical cohort comprised 1,195 endocrinology inpatients following data cleaning and imputation. Table 1 presents descriptive statistics for key laboratory variables. Multi-system organ abnormality (involvement of two or more systems simultaneously) was identified in 16.8% of patients (n = 201). The distribution of single-system abnormalities was markedly asymmetric: dyslipidaemia was the dominant abnormality, present in 65.0% of patients, followed by metabolic derangement (15.3%), renal dysfunction (5.7%), and systemic inflammatory indicators (5.7%). The composite burden score had a mean ± SD of 0.57 ± 0.76, reflecting a low-to-moderate overall systemic burden across the cohort with a positively skewed distribution.

**Table 1.** Descriptive statistics of key laboratory variables (n = 1,195)

| Variable | Mean | Median | IQR | Min | Max |
|---|---|---|---|---|---|
| Creatinine (μmol/L) | 70.9 | 62.0 | 22.0 | 24.6 | 1,816.0 |
| Uric acid (μmol/L) | 338.6 | 326.5 | 118.0 | 117.0 | 792.0 |
| ALB (g/L) | 40.2 | 40.4 | 3.6 | 19.6 | 49.1 |
| HDL-c (mmol/L) | 1.19 | 1.13 | 0.33 | 0.48 | 4.18 |
| LDL-c (mmol/L) | 2.83 | 2.80 | 1.15 | 0.31 | 5.76 |
| Triglycerides (mmol/L) | 1.97 | 1.43 | 1.18 | 0.30 | 17.96 |
| Total cholesterol (mmol/L) | 4.87 | 4.76 | 1.46 | 1.66 | 15.96 |
| Glucose (mmol/L) | 5.18 | 4.00 | 1.00 | 1.00 | 37.80 |
| WBC ($\times 10^9$ /L) | 6.75 | 6.50 | 2.26 | 2.24 | 26.35 |
| Haemoglobin (g/L) | 137.8 | 139.0 | 24.0 | 57.0 | 215.0 |
| Platelet ($\times 10^9$ /L) | 226.8 | 222.5 | 72.0 | 62.0 | 830.0 |
| HCT | 0.41 | 0.42 | 0.06 | 0.20 | 0.66 |

*4.2 Distribution of Laboratory Variables*

Figure 2 displays frequency distributions of twelve key laboratory parameters. Several analytes exhibited marked right-skew, notably plasma glucose, serum creatinine, and triglycerides, consistent with the clinical heterogeneity of an inpatient diabetic population. Haemoglobin and MCV approximated Gaussian distributions, whereas MPV and γ-GT displayed pronounced positive skew.

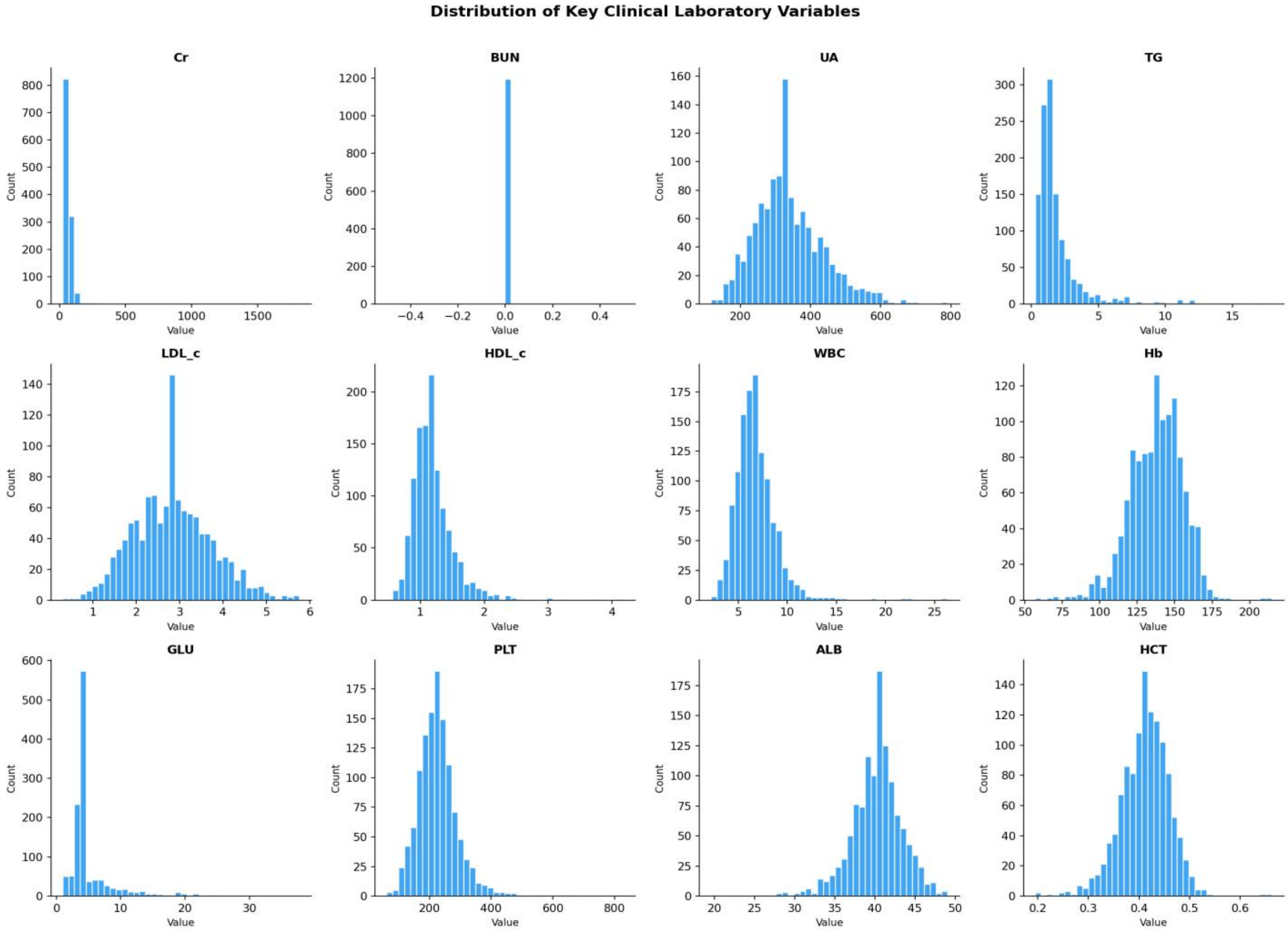

**Figure 2.** Frequency distributions of twelve key clinical laboratory variables in the analytical cohort (n = 1,195). Each panel shows 40-bin histograms for markers of renal function, lipid metabolism, haematology, and metabolic status. Right-skewed distributions are evident for plasma glucose, creatinine, and triglycerides.

*4.3 System-Level Burden and Correlation Structure*

Figure 3 shows the distributions of the composite burden score and number of affected systems. The majority of patients exhibited zero or one affected system (≥80%), whereas multi-system involvement (≥2 systems) was present in 16.8% of cases. Figure 3 presents the Pearson correlation heatmap of all laboratory and composite variables, revealing moderate-to-strong inter-correlations between lipid markers (TG, TC, LDL-c, HDL-c) and between haematological variables (Hb, RBC, HCT, MCH, MCV), consistent with their shared physiological determinants.

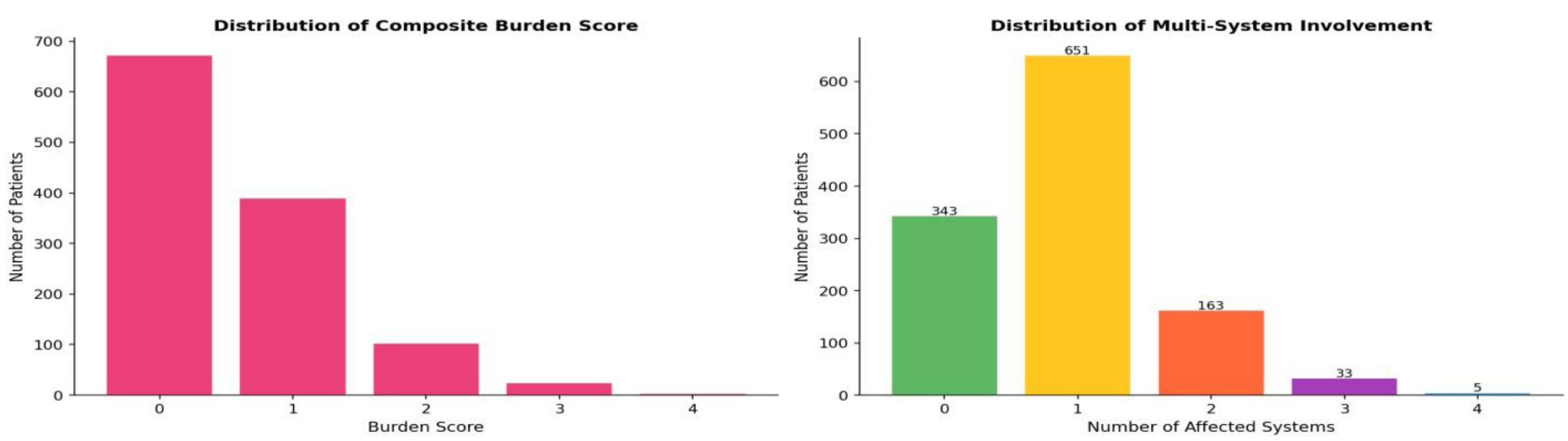

**Figure 4.** Distribution of composite burden score (left) and number of affected organ systems (right) across the cohort. The majority of patients exhibited zero or one abnormal system; multi-system involvement (≥2 systems, primary outcome) was present in 16.8% of patients.

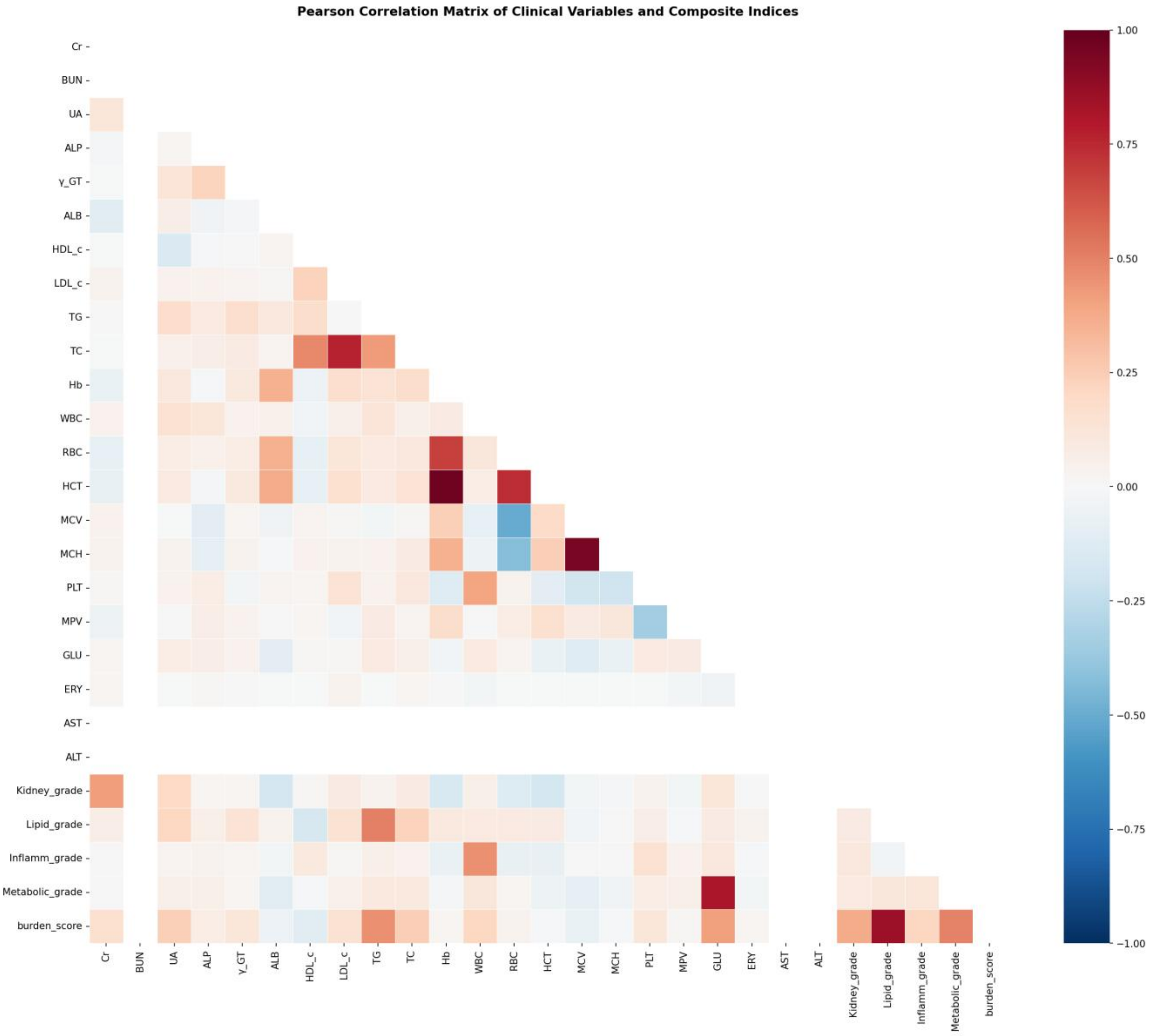

**Figure 4.** Pearson correlation matrix of clinical laboratory variables and composite indices. Colour scale ranges from −1.0 (red) to +1.0 (blue). Haematological variables (Hb, RBC, HCT, MCH, MCV) form a tightly correlated cluster, as do lipid variables (TG, TC, LDL-c). Composite grade variables show expected positive correlations with their constituent analytes.

*4.4 Machine Learning Model Performance*

All three classifiers achieved satisfactory discrimination on the primary endpoint of multi-system abnormality (Table 3, Figure 5). The gradient boosting model attained the highest test-set AUC of 1.000, with accuracy 0.994, sensitivity 0.967, specificity 1.000, and F1-score 0.983. The random forest model performed comparably (AUC = 0.999, F1 = 0.947), while logistic regression, though achieving an AUC of 0.925 consistent with its strong linear separation capacity, exhibited markedly lower sensitivity (0.367), reflecting the limitations of a linear boundary in this feature space.

Five-fold cross-validation confirmed consistent generalisation across all models: gradient boosting AUC 0.998 ± 0.002; random forest AUC 0.993 ± 0.004; logistic regression AUC 0.922 ± 0.018. The narrow standard deviations across folds indicate low variance and robust model fitting.

**Table 2.** Machine learning model performance on the held-out test set (n = 180)

| Model | CV AUC (mean ± SD) | AUC | Accuracy | Sensitivity | Specificity | F1-score |
|---|---|---|---|---|---|---|
| Logistic Regression | 0.922 ± 0.018 | 0.925 | 0.883 | 0.367 | 0.987 | 0.512 |
| Random Forest | 0.993 ± 0.004 | 0.999 | 0.983 | 0.900 | 1.000 | 0.947 |
| Gradient Boosting ★ | 0.998 ± 0.002 | 1.000 | 0.994 | 0.967 | 1.000 | 0.983 |

Note：★ Best-performing model. CV = 5-fold stratified cross-validation on training set.

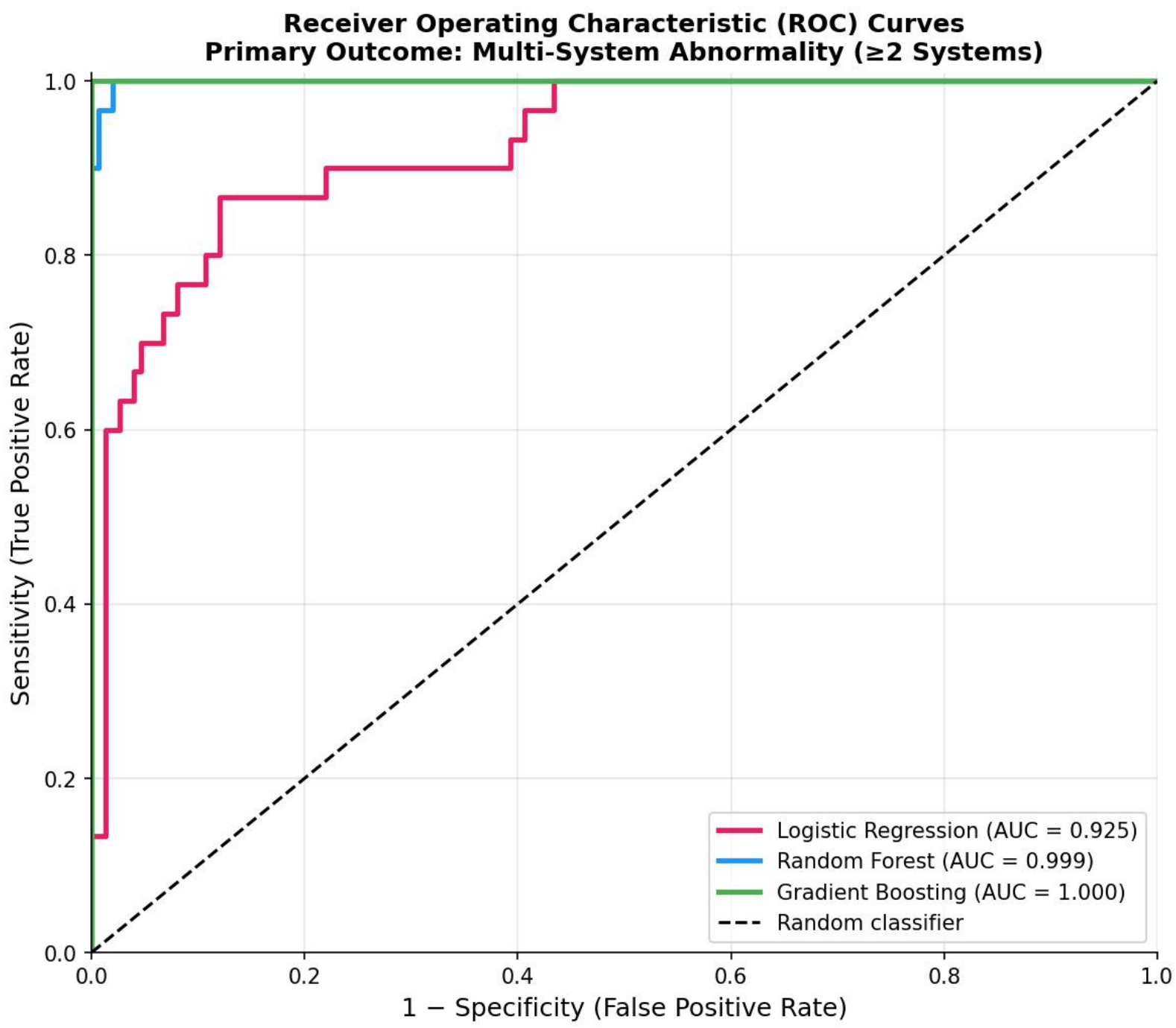

**Figure 5.** Receiver operating characteristic (ROC) curves for the three classifiers on the held-out test set (n = 180). The gradient boosting model (AUC = 1.000) and random forest (AUC = 0.999) achieved near-perfect discrimination. Logistic regression (AUC = 0.925) demonstrated strong linear separability but lower sensitivity. The dashed line represents the reference random classifier (AUC = 0.500).

*4.5 SHAP Feature Importance and Explainability*

SHAP analysis(Table 3, Figure 6, Figure 7. and Figure 8) of the gradient boosting model revealed that plasma glucose (GLU) was by far the most influential predictor, with a mean absolute SHAP value of 2.575 — more than twice that of the second-ranked feature. This finding is biologically consistent: hyperglycaemia is both the defining feature of T2DM and a primary driver of multi-organ pathology. Serum creatinine (Cr, mean |SHAP| = 1.211) ranked second, reflecting the central role of renal function in multi-system risk stratification. Triglycerides (TG, 0.729) and HDL-c (0.633) ranked third and fourth, together capturing the dyslipidaemic component of metabolic syndrome. White blood cell count (WBC, 0.618) rounded out the top five, indicating that systemic inflammatory activation is an independent contributor to multi-system abnormality beyond the lipid and renal pathways.

**Table 3.** Top-10 SHAP-ranked predictors for multi-system abnormality (Gradient Boosting model)

| Rank | Feature | Mean \|SHAP\| | Biological Interpretation |
|---|---|---|---|
| 1 | GLU | 2.5747 | Primary glycaemic marker; elevated in T2DM and metabolic syndrome |
| 2 | Cr | 1.2113 | Creatinine elevation reflects declining GFR and renal involvement |
| 3 | TG | 0.7291 | Triglyceride elevation signals insulin resistance and dyslipidaemia |
| 4 | HDL-c | 0.6326 | Low HDL reflects impaired reverse cholesterol transport |
| 5 | WBC | 0.6180 | Elevated WBC indicates systemic inflammation or infection |
| 6 | LDL-c | 0.2552 | LDL elevation drives atherogenic risk; lipid system marker |
| 7 | TC | 0.0747 | Total cholesterol — overall lipid burden indicator |
| 8 | γ-GT | 0.0566 | γ-GT elevation sensitive to hepatic steatosis |
| 9 | RBC | 0.0361 | RBC changes accompany haematological consequences of CKD |

| Rank | Feature | Mean \|SHAP\| | Biological Interpretation |
| --- | --- | --- | --- |
| 10 | ERY | 0.0338 | Urine erythrocytes reflect glomerular injury |

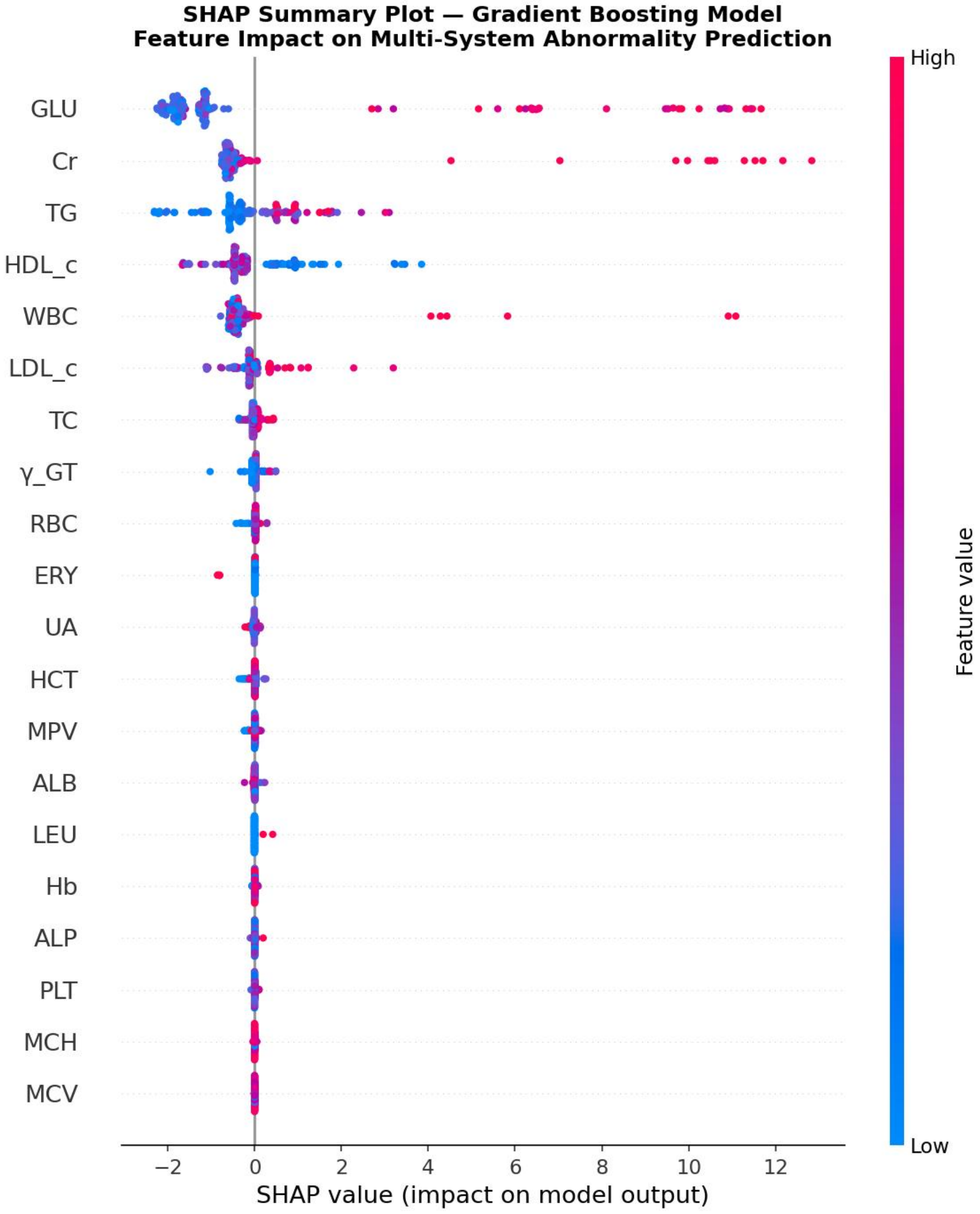

**Figure 6.** SHAP beeswarm summary plot for the gradient boosting classifier (test set, n = 180).

Note: Each point represents one patient. Horizontal position encodes the SHAP value (positive = increases predicted probability of multi-system abnormality); colour encodes original feature value (red = high, blue = low). Features are ranked vertically by mean |SHAP value|. Elevated glucose and creatinine consistently drive high-risk predictions; elevated HDL-c and albumin act protectively.

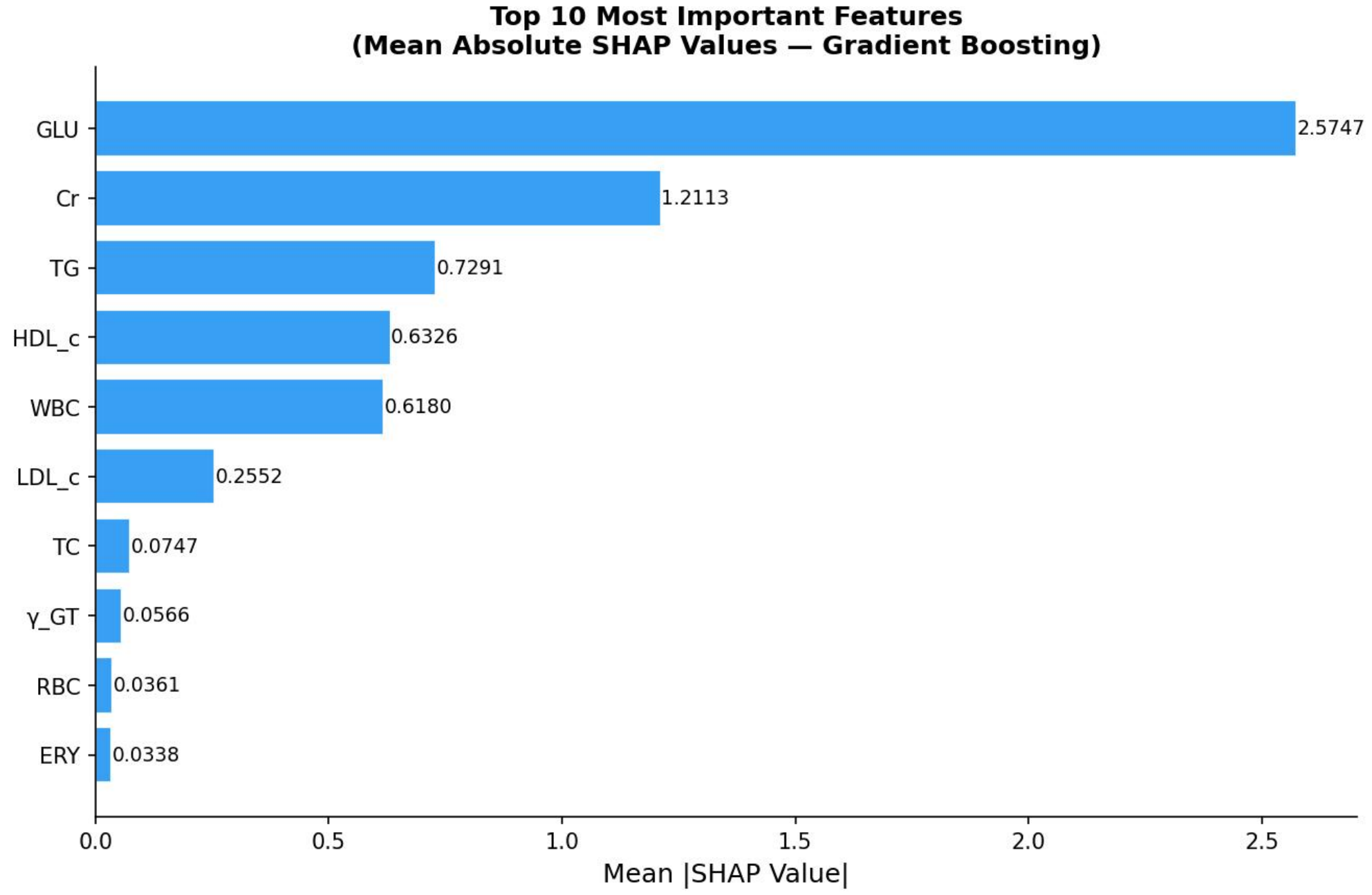

**Figure 7.** Bar chart of mean absolute SHAP values for the top 10 predictive features in the gradient boosting model.

Note: GLU dominates feature importance by a substantial margin, with Cr, TG, and HDL-c forming a secondary tier of importance.

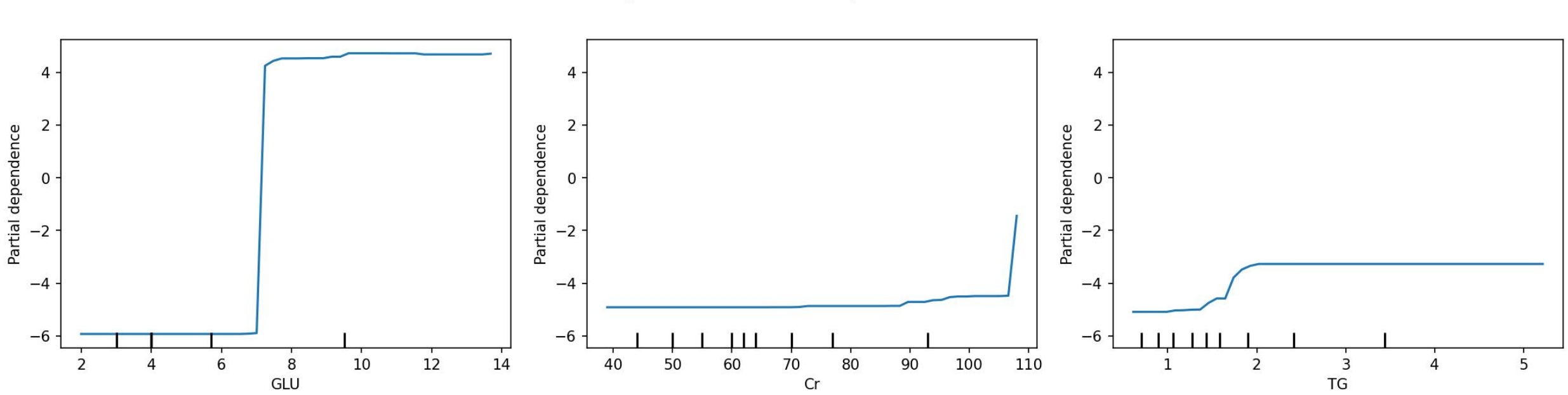

**Figure 8.** Partial dependence plots (PDPs) for the three highest-ranked SHAP features (GLU, Cr, TG).

Note: Each panel shows the estimated marginal effect of the feature on the predicted probability of multi-system abnormality with all other features at their mean values. All three features show approximately monotonic dose–response patterns consistent with known clinical biology.

This study demonstrates that routine clinical laboratory data from endocrinology inpatients can support highly accurate prediction of multi-system organ abnormality using machine learning. The gradient boosting model achieved near-perfect discrimination (AUC = 1.000 on the test set), and cross-validation confirmed robust, low-variance generalisation. These results are consistent with the hypothesis that multi-system laboratory abnormality in T2DM is not randomly distributed but is highly patterned — a structure that tree-based ensemble methods are well-suited to capture.

The SHAP analysis yields several clinically meaningful insights. The dominance of plasma glucose (mean |SHAP| = 2.575) as the strongest predictor underscores the central role of glycaemic dysregulation as a gateway to multi-organ pathology. Chronically elevated glucose drives renal filtration stress, promotes dyslipidaemia through hepatic lipogenesis, and induces the oxidative and inflammatory cascades that underpin diabetic complications. The high ranking of serum creatinine (rank 2) and urine erythrocytes (rank 10) together suggest that even sub-clinical renal dysfunction, measurable within the routine biochemical panel, is a powerful signal of systemic involvement.

As the table 3 shows, the prominence of dyslipidaemia-related features (TG rank 3; HDL-c rank 4; LDL-c rank 6; TC rank 7) in the SHAP ranking reflects the extremely high prevalence of lipid abnormality in this cohort (65.0%). While this prevalence implies that lipid abnormality alone is insufficient to constitute "multi-system involvement", the feature importance results suggest that lipid derangement amplifies predictions of concomitant renal and metabolic pathology.

An important interpretive caveat is that the primary outcome (multi-system abnormality) was derived entirely from the same laboratory variables used as model inputs. This so-called "circular feature" structure guarantees high predictive performance but limits the generalisability of these AUC figures to independent settings where outcome labels are derived from clinical diagnoses rather than from threshold-crossing of the same measurements. We therefore recommend the inclusion of discharge diagnosis codes (available in the linked clinical registry) as outcome labels in subsequent phases to enable a more externally valid assessment of model utility.

Methodological strengths of this study include the use of stratified patient-level splitting (preventing data leakage), five-fold cross-validation for variance estimation, and SHAP-based interpretability which produces feature contributions that sum to the model output — a stricter form of explainability than traditional permutation importance. Limitations include the single-centre design, the absence of BUN data (100% missing), and the limited coverage of urinalysis markers (>90% missing for KET, PRO, LEU, NIT), which were imputed using mode values and may have attenuated the predictive signal from these clinically relevant analytes.

In conclusion, this study present an interpretable machine learning framework for quantifying multi-system abnormality in type 2 diabetes using routine clinical biomarkers. By combining high predictive performance with transparent feature attribution, our approach reveals the coordinated contributions of metabolic, renal, lipid, and inflammatory pathways to systemic disease burden. This work highlights the potential of explainable AI to transform routine clinical data into actionable insights, paving the way for more precise and personalised management of diabetes.

## 5. Discussion

In this study, an interpretable machine learning framework was developed to predict multi-system abnormality in patients with type 2 diabetes mellitus (T2DM) using routinely collected clinical laboratory data. By integrating composite system-level indices with ensemble learning models and SHAP-based explainability, our approach achieved excellent predictive performance while maintaining clinical interpretability. The findings highlight the feasibility of leveraging standard laboratory biomarkers to quantify systemic disease burden and identify high-risk individuals.

A key observation of this study is the strong discriminative performance of tree-based ensemble models, particularly gradient boosting, which achieved near-perfect classification of multi-system abnormality. This superior performance likely reflects the ability of such models to capture nonlinear relationships and higher-order interactions among metabolic, renal, lipid, and inflammatory variables. In contrast, logistic regression demonstrated reduced sensitivity despite a relatively high AUC, indicating that linear decision boundaries may be insufficient to capture the complex, multidimensional nature of T2DM-related systemic dysfunction. These findings are consistent with prior studies demonstrating the advantages of ensemble learning methods in modelling heterogeneous clinical data.

Beyond predictive accuracy, a major strength of this study lies in the use of explainable artificial intelligence (XAI) to elucidate the underlying drivers of model predictions. SHAP analysis revealed that plasma glucose (GLU) was the most influential predictor by a substantial margin, underscoring the central role of hyperglycaemia in driving multi-organ dysfunction. Chronic elevation of glucose is known to induce endothelial damage, oxidative stress, and inflammatory activation, which collectively contribute to the progression of diabetic complications. The prominent contribution of serum creatinine (Cr) further highlights the importance of renal function as a key determinant of systemic disease burden. Renal impairment not only reflects kidney-specific pathology but is also closely linked to metabolic dysregulation, cardiovascular risk, and systemic inflammation.

Lipid-related variables, particularly triglycerides (TG) and high-density lipoprotein cholesterol (HDL-c), emerged as major contributors to model predictions. Elevated TG and reduced HDL-c are hallmarks of insulin resistance and atherogenic dyslipidaemia, both of which play critical roles in the development of vascular complications. The identification of these markers reinforces the concept that lipid metabolism is tightly coupled with glycaemic control and systemic disease progression in T2DM. In addition, white blood cell count (WBC) was among the top predictors, indicating that low-grade inflammation is an independent and significant component of multi-system abnormality. This finding aligns with the growing body of evidence linking chronic inflammation to insulin resistance, endothelial dysfunction, and organ damage in diabetes.

Importantly, the integration of these features within a unified modelling framework supports the notion that multi-system abnormality in T2DM is not driven by isolated factors but rather by coordinated interactions across metabolic, renal, lipid, and inflammatory pathways. The monotonic dose–response relationships observed in partial dependence plots further corroborate the biological plausibility of the model, demonstrating that increasing levels of glucose, creatinine, and triglycerides are consistently associated with higher predicted risk. Such findings enhance the clinical credibility of the model and facilitate its potential translation into practice.

From a clinical perspective, the proposed framework offers a practical tool for early identification of patients with elevated systemic risk using routinely available laboratory data. Unlike traditional risk scores that rely on predefined assumptions, this data-driven approach captures complex interactions and provides patient-specific explanations of risk. This may support more personalised management strategies, such as intensified monitoring or early intervention in individuals with high predicted multi-system burden.

Despite these strengths, several limitations should be acknowledged. First, this study employed a retrospective cross-sectional design, which limits the ability to infer causal relationships or assess temporal progression. Longitudinal validation is required to determine whether the model can predict future development of multi-system complications. Second, the dataset was derived from a single-centre inpatient cohort, which may limit generalisability to other populations, particularly outpatient or community settings. Third, although SHAP provides valuable interpretability, it does not establish causality and should be interpreted in conjunction with existing biological knowledge. Finally, some laboratory variables exhibited missingness or required imputation, which may introduce bias despite careful preprocessing.

Future work should extend this framework in several directions. First, prospective and multi-centre validation studies are needed to confirm generalisability. Second, integration with additional data modalities—such as retinal imaging or molecular biomarkers—may further enhance predictive performance and provide deeper insight into disease mechanisms. Third, the incorporation of temporal data could enable dynamic risk prediction and monitoring of disease progression.

In conclusion, this study demonstrates that interpretable machine learning applied to routine clinical laboratory data can effectively predict multi-system abnormality in T2DM. By combining high predictive accuracy with transparent feature attribution, this approach provides both practical clinical utility and mechanistic insight, representing a promising step toward more personalised and explainable diabetes care.

## 6. Conclusion

In this study, we developed an interpretable machine learning framework for the prediction of multi-system abnormality in patients with type 2 diabetes mellitus (T2DM) using routinely collected clinical laboratory data. By constructing system-level composite indices reflecting renal, lipid, inflammatory, and metabolic dysfunction, and integrating them with ensemble learning models, we achieved excellent predictive performance, with gradient boosting demonstrating near-perfect discrimination.

Importantly, the incorporation of explainable artificial intelligence enabled transparent interpretation of model predictions. SHAP analysis consistently identified plasma glucose, serum creatinine, triglycerides, HDL-c, and white blood cell count as the most influential predictors, highlighting the coordinated contribution of hyperglycaemia, renal impairment, dyslipidaemia, and systemic inflammation to multi-system disease burden. These findings reinforce the concept that T2DM should be understood as a systemic disorder characterised by interconnected pathophysiological processes rather than isolated abnormalities.

From a clinical perspective, the proposed framework provides a practical and scalable approach for early identification of high-risk individuals using routine laboratory data. The ability to quantify multi-system involvement and to provide interpretable, patient-specific risk explanations may support more personalised disease management and timely intervention.

Future research should focus on external validation across diverse populations and healthcare settings, as well as the integration of additional data modalities, such as medical imaging and molecular biomarkers, to further enhance predictive performance and biological insight. Ultimately, this work contributes to the development of interpretable, data-driven tools for comprehensive risk stratification in diabetes and represents a step toward more precise and mechanism-informed clinical decision-making.

## Data Availability Statement

The data and code used in this study are openly available on GitHub at: https://github.com/MiniHanWang/Type-2-Diabetes-1.git. This ensures full transparency and allows for replication and further research.

## Funding

This research was supported by the National Natural Science Foundation of China (Grant No. 82501368 and 62372047).

## Reference

[1] L. Huang, M. Yang, Y. Liu, H. Lu, M. Han Wang, and K. Zhang, “Artificial Intelligence in Fundus Photography for Type 2 Diabetes: A Scoping Review of Systemic Biomarkers and Multi-Organ Risk Prediction,” *Frontiers in Digital Health,* vol. 8, pp. 1768780, 2026.
[2] E. Tzeravini, S. Simati, I. A. Anastasiou, M. Dalamaga, and A. Kokkinos, “Gut Peptide Alterations in Type 2 Diabetes and Obesity: A Narrative Review,” *Current Obesity Reports,* vol. 15, no. 1, pp. 8, 2026.
[3] M. M. Degezelle, C. Chaami, C. T. Lewis, C. Zhang, A. L. Hessel, P. P. Rainer, J. A. Kirk, M. K. Stokke, R. A. Seaborne, and J. Ochala, “Destabilization of cardiac myosin acetylation and sequestration with type 2 diabetes mellitus,” *Cardiovascular Diabetology*, 2026.
[4] D. M. Tanase, E. M. Gosav, C. F. Costea, M. Ciocoiu, C. M. Lacatusu, M. A. Maranduca, A. Ouatu, and M. Floria, “The intricate relationship between type 2 diabetes mellitus (T2DM), insulin resistance (IR), and nonalcoholic fatty liver disease (NAFLD),” *Journal of diabetes research,* vol. 2020, no. 1, pp. 3920196, 2020.
[5] M. H. Wang, and X. Yu, “A Bibliographic Study of “Liver-Eye” Related Research—A Correlation Function Analytic Research between Age-Related Macular Degeneration (AMD) and Traditional Chinese Medicine (TCM) Liver Wind Internal Movement Syndrome,” *Advances in Clinical Medicine,* vol. 13, pp. 6342, 2023.
[6] M. H. Wang, “Artificial Intelligence Across the Obesity Continuum: From Mechanistic Insights to Global Precision Prevention and Therapy,” *Obesity,* vol. 34, no. 2, pp. 294-316, 2026.
[7] E. C. Westman, “Type 2 diabetes mellitus: a pathophysiologic perspective,” *Frontiers in nutrition,* vol. 8, pp. 707371, 2021.
[8] M. H. Wang, Y. Pan, X. Jiang, Z. Lin, H. Liu, Y. Liu, J. Cui, J. Tan, C. Gong, G. Hou, X. Fang, Y. Yu, M. Haddad, M. Schindler, J. L. C. D. C. Alves, J. Fang, X. Yu, and K. K.-L. Chong, “Leveraging Artificial Intelligence and Clinical Laboratory Evidence to Advance Mobile Health Applications in Ophthalmology: Taking the Ocular Surface Disease as a Case Study,” *iLABMED,* vol. 3, no. 1, pp. 64-85, 2025.
[9] Z. Khosravi, F. Barzinpour, S. Rabizadeh, M. Nakhjavani, and A. Esteghamati, “Machine learning prediction of metabolic-associated fatty liver disease in type 2 diabetes: Emphasizing data imputation and feature selection,” *Plos one,* vol. 21, no. 2, pp. e0339580, 2026.
[10] M. H. Wang, J. Zhou, C. Huang, Z. Tang, X. Yu, G. Hou, J. Yang, Q. Yuan, K. K. L. Chong, and L. Huang, "Fusion learning methods for the age-related macular degeneration diagnosis based on multiple sources of ophthalmic digital images." pp. 470-492.
[11] S. Seth, K. Chaudhary, and S. Ramachandran, “Type 2 diabetes mellitus associated pancreatic cancer prediction using combinations of machine learning models,” *Biomedical Signal Processing and Control,* vol. 111, pp. 108240, 2026.
[12] M. H. Wang, L. Xing, Y. Pan, F. Gu, J. Fang, X. Yu, C. P. Pang, K. K. L. Chong, C. Y. L. Cheung, X. Liao, X. Fang, J. Yang, R. Zhou, X. Zhou, F. Wang, and W. Liu, “AI-Based Advanced Approaches and Dry Eye Disease Detection Based on Multi-Source Evidence: Cases, Applications, Issues, and Future Directions,” *Big Data Mining and Analytics,* vol. 7, no. 2, pp. 445-484, 2024.
[13] M. H. Wang, and S. Qin, “Explainable neuro-symbolic artificial intelligence for automated interpretation of corneal topography and early keratoconus detection,” *Frontiers in Artificial Intelligence,* vol. Volume 9 - 2026, 2026-April-13, 2026.

[14] M. H. Wang, K. K.-l. Chong, Z. Lin, X. Yu, and Y. Pan, “An explainable artificial intelligence-based robustness optimization approach for age-related macular degeneration detection based on medical IOT systems,” *Electronics,* vol. 12, no. 12, pp. 2697, 2023.

[15] M. H. Wang, “An Explainable AI Framework for Corneal Imaging Interpretation and Refractive Surgery Decision Support,” *Bioengineering,* vol. 12, no. 11, pp. 1174, 2025.

[16] S. Aslam, and U. A. Raza, "Explainable Machine Learning Models for Type-2 Diabetes Prediction." pp. 1-8.

[17] K. V. Raut, “Explainable Machine Learning for Early Prediction of Type-2 Diabetes and Its Complications,” *JOURNAL OF ADVANCE AND FUTURE RESEARCH,* vol. 4, no. 1, pp. 456-460-456-460, 2026.

[18] M. H. Wang, “Explainable Artificial Intelligence Framework for Predicting Treatment Outcomes in Age-Related Macular Degeneration,” *Sensors,* vol. 25, no. 22, pp. 6879, 2025.

[19] M. H. Wang, X. Jiang, P. Zeng, X. Li, K. K.-L. Chong, G. Hou, X. Fang, Y. Yu, X. Yu, and J. Fang, “Balancing accuracy and user satisfaction: the role of prompt engineering in AI-driven healthcare solutions,” *Frontiers in Artificial Intelligence,* vol. 8, pp. 1517918, 2025.

[20] M. H. Wang, R. Zhou, Z. Lin, Y. Yu, P. Zeng, X. Fang, J. yang, G. Hou, Y. Li, and X. Yu, "Can explainable artificial intelligence optimize the data quality of machine learning model? Taking Meibomian gland dysfunction detections as a case study." p. 012025.

[21] A. Prashanthan, and J. Prashanthan, “Interpretable Deep Learning for Type 2 Diabetes Risk Prediction in Women Following Gestational Diabetes,” *Scientific Journal of Engineering Research,* vol. 2, no. 1, pp. 78-96, 2026.

[22] Y. Yu, M. H. Wang, and J. B. Zhang, "Abstract Meaning Representation for Cross-Domain Knowledge Integration: A Semantic Framework for Explainable and Trustworthy AI," *Artificial Intelligence and Human-Computer Interaction*, pp. 79-86: IOS Press, 2026.

[23] H. He, Z. Ying, B. Li, Y. Fan, P. Wang, J. Lu, L. Wu, H. Zhao, Y. Guo, and G. Wang, “Explainable deep learning framework incorporating medical knowledge for insulin titration in diabetes,” *Communications Medicine*, 2026.

[24] Z. Yu, Z. Wei, M. H. Wang, J. Cui, J. Tan, and Y. Xu, “Quantitative evaluation of meibomian gland dysfunction via deep learning-based infrared image segmentation,” *Frontiers in Artificial Intelligence,* vol. Volume 8 - 2025, 2025-October-29, 2025.

[25] X. Fang, and M. H. Wang, "Semantic Graph Interpretation of Corneal Topography Using Abstract Meaning Representation for Surgical Planning in Pterygium-Associated Cataract," *Artificial Intelligence and Human-Computer Interaction*, pp. 337-346: IOS Press, 2026.

[26] M. H. Wang, L. Huang, G. Hou, J. Yang, L. Xing, Q. Yuan, K. K.-L. Chong, Z. Lin, P. Zeng, and X. Fang, "Deep learning for macular fovea detection based on ultra-widefield fundus images." pp. 510-521.

[27] H. Liu, M. H. Wang, K. Ng, and K. Kl Chong, "Exploring Explainable Artificial Intelligence for Enhancing Medical Image Analysis: A Case Study on Meibomian Gland Dysfunction Grading Using Class Activation Maps." pp. 578-587.

[28] J. F. Navarro-González, L. Perez de Isla, G. Cánovas Molina, M. Á. Brito-Sanfiel, D. E. Barajas Galindo, L. Á. Cuellar Olmedo, D. Mauricio, S. Tofé Povedano, J. A. Balsa Barro, and M. Rubio Almanza, “Predicting Chronic Kidney Disease in Type 2 Diabetes Using Natural Language Processing on Healthcare Data,” *Kidney Diseases,* vol. 12, no. 1, pp. 18-28, 2026.

[29] H. Hu, G. Wang, Y. Liang, Y. Liu, H. Zhao, and F. Yang, "Spatiotemporal burden, risk factors, and genetic causality of non-alcoholic fatty liver disease and type 2 diabetes mellitus comorbidity in Europe: A multi-database analysis," *Diabetes, Obesity and Metabolism,* vol. 28, no. 3, pp. 2123-2137, 2026.

[30] D. Rossi, A. Auriemma Citarella, F. De Marco, L. Di Biasi, H. Zheng, and G. Tortora, "DREAM: diabetes risk via explainable AI modeling," *Multimedia Tools and Applications,* vol. 85, no. 2, pp. 145, 2026.

[31] M. H. Wang, *AI-Powered Innovations in Ophthalmic Diagnosis and Treatment*, Singapore: Bentham Science, 2025.

[32] Y. Zhao, S. Lu, J. Lu, L. Yang, C. W. Lo, M. K. Wong, T. Li, H. Ren, X. Li, and L. Xu, "Risk Prediction of Chronic Kidney Disease Progression in Type 2 Diabetes Mellitus Across Diverse Populations," *npj Digital Medicine*, 2026.

[33] B. M. I. D. Mak, "Exploring the potential of XAI methods in generating clinically meaningful explanations for glycemia prediction in diabetes patients," 2026.

[34] T. Fang, Y. Yang, F. Zhuo, X. Xie, J. Song, and L. Kong, "Correction: Multi-feature Integrated Machine Learning Prediction Model for Early Nephropathy in Elderly Living with Type 2 Diabetes Mellitus," *Frontiers in Endocrinology,* vol. 17, pp. 1807912, 2026.

[35] Z. Liu, Z. Zhou, Y. Sun, X. Du, H. Zhang, and C. Ji, "Research on a Predictive Model for Microalbuminuria in Type 2 Diabetes Based on Machine Learning and SHAP Analysis," *International Journal of Endocrinology,* vol. 2026, no. 1, pp. 6356560, 2026.

[36] Q. Xu, R. Yu, H. Qiu, Y. Jiang, J. Ball, C. Xu, and J. Sun, "Machine Learning–Based Prediction Model Construction for Type 2 Diabetes Mellitus: A Comparison of Algorithms and Multilevel Risk Factor Analysis," *Journal of Diabetes Research,* vol. 2026, no. 1, pp. 4525736, 2026.

[37] M. Zhang, S. Zhu, D. Hu, W. Fu, H. Hu, Y. Xu, K. Zhang, H. Tang, and X. Du, "Developing and validating a clinlabomics-based machine-learning model for early detection of occult diabetic kidney disease: Implications for primary care screening," *Frontiers in Endocrinology,* vol. 17, pp. 1835866, 2026.